\documentclass[conference]{IEEEtran}
\IEEEoverridecommandlockouts

\usepackage{cite}
\usepackage{amsmath,amssymb,amsfonts}
\usepackage{algorithmic}
\usepackage{graphicx}
\usepackage{textcomp}
\usepackage{multirow}
\usepackage{multicol}

\usepackage{multirow,tabularx}
\usepackage{textcomp}
\usepackage{xcolor}
\usepackage{amsmath}
\usepackage{amsfonts}
\usepackage[T1]{fontenc}
\usepackage{float}
\usepackage{url}
\usepackage{balance}

\usepackage{xcolor}
\def\BibTeX{{\rm B\kern-.05em{\sc i\kern-.025em b}\kern-.08em
    T\kern-.1667em\lower.7ex\hbox{E}\kern-.125emX}}
\begin{document}
\title{CLAP: A Convolutional Lightweight Autoencoder for Plant Disease Classification \\
{
}
}
\author{\IEEEauthorblockN{Asish Bera$^{*}$,~\IEEEmembership{Senior Member,~IEEE}, Subhajit Roy, and  Sudiptendu Banerjee }
\IEEEauthorblockA{\\ Department of Computer Science and Information Systems, Birla Institute of Technology and Science Pilani, Rajasthan, India\\  \texttt{\{asish.bera,  subhajit.roy, sudiptendu.banerjee\}@pilani.bits-pilani.ac.in} \\
}
\thanks{This work was supported by the Prime Minister Early Career Research Grant (PMECRG 2024), ANRF, File No: ANRF/ECRG/2024/001743/ENS}
\thanks{$^*$ Corresponding Author: A. Bera  }
%
}

\maketitle

\begin{abstract}
 Convolutional neural networks (CNNs) have remarkably progressed the performance of distinguishing  plant diseases, severity grading, and nutrition deficiency prediction using leaf images. However, these tasks become more challenging in a realistic in-situ field  condition. Often, a traditional machine learning model may fail to capture and interpret discriminative characteristics of plant health, growth and diseases due to subtle variations within leaf sub-categories. A few deep learning methods have used additional preprocessing stages or network modules to address the problem, whereas several  other methods have utilized pre-trained backbone CNNs, most of which are computationally intensive. Therefore, to address the challenge, we propose a lightweight autoencoder using separable convolutional layers in its encoder-decoder blocks. A sigmoid-gating is applied for refining the prowess of the encoder's feature discriminability, which is improved further by the decoder. Finally, the feature maps of encoder-decoder are combined for rich feature representation before classification. The proposed Convolutional Lightweight Autoencoder for Plant disease classification, called CLAP, has been experimented on three public plant datasets consisting of cassava, tomato, maize, groundnut, grapes, etc. for determining plant health conditions.
The CLAP has attained improved or competitive accuracies  on the Integrated Plant Disease (95.67\%), Groundnut (96.85\%), and  CCMT (87.11\%) datasets balancing a trade-off between the performance, and little computational  cost requiring  5 million parameters. The training time is 20 milliseconds (ms) and inference time is 1 ms per image. 
 
\end{abstract}


\section{Introduction}
Agriculture serves a vital role  for sustainable economic  growth of a nation. Improvement of crop production  is crucial to meet  global food demands. However, crop yield and quality control is a major concern  due to adverse environmental conditions, climate change, global warming,  and insufficient  infrastructure  \cite{kabato2025towards}. Also,  proper soil moisture, micro nutrients, and water  management are equally important for plant growth. 
So, scientific nourishment and monitoring of plant health and stress management is a demanding task for improving crop yield. 
Recent technological development of agriculture has witnessed the impacts of  artificial intelligence (AI) and  machine learning (ML) techniques over manual supervision of the farmers and agronomist.  Automated agricultural systems empowered with AI-ML  are very effective to develop accurate predictive models to deal with plant stress at early stages.

\begin{figure} [h]
\includegraphics[width=0.48\textwidth]{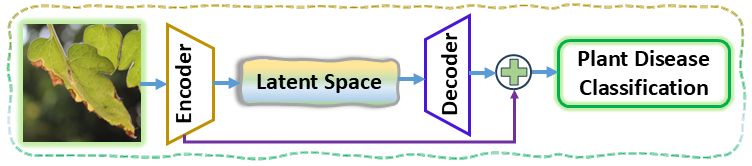} 
\caption{CLAP: convolutional lightweight autoencoder to classify plant stress   
}
\label{CLAP}
\end{figure}

Pertaining to precision agriculture, several challenges are investigated which need more scientific attention \textit{e.g.},  early disease detection and classification, nutrition deficiency identification, and others \cite{goyal2025custom},  \cite{bera2024pnd}.  %
Classification of plant disease by analyzing visual symptoms in leaf images  is one of the  significant intricate tasks. Plant disease depicts a broader  and subtle visual variations in leaf images, which is cumbersome due to lighting and field conditions.  
To address these challenges, deep learning (DL) models, especially convolutional neural networks (CNNs) and others have evinced their efficacy. In this direction, advanced computer vision based DL techniques, language models, and more have recently been explored.

Several ML/DL methods  for plant disease prediction have been developed. Sample images of plants are acquired and  pre-processed with different  methods \textit{e.g.}, noise removal, area localization, boundary extraction, etc.
  Also, lightweight CNNs and autoencoders requiring less computational parameters are suitable for plant health and disease detection \cite{chen2022dfcanet}. 

In this work, a \textbf{C}onvolutional \textbf{L}ightweight \textbf{A}utoencoder for \textbf{P}lant disease classification, named CLAP, is proposed.  {Many existing works used pre-trained CNN or vision transformer (ViT) backbones which are computationally moderate to heavier models. Recent works develop lightweight model architecture for improving performance at a lower computational overhead, \textit{e.g.}, lightweight ViTs are very effective. Though convolutional auto-encoder (CAE) achieves competitive performances, however, CAEs have received a little research attention. Thus, we propose a simple  convolutional lightweight auto-encoder (CLA) that offers a balance between the performance and computational complexity.}
The CLAP tested on multiple plant datasets has achieved improved disease classification performances compared to existing methods. Notable contributions of this work are:

\begin{itemize}
\item A lightweight convolutional autoencoder (CLAP) using depthwise separable convolutional layers is  proposed  for plant disease classification. 

\item The proposed CLAP is effective for improving the performance on three  public datasets consisting of leaf images of various plants, implying its generalization ability. 
\end{itemize}

The rest of this paper is organized as follows:  related works are briefed in Section \ref{Rwrk}. The proposed method is  described in Section \ref{method}.  The  experimental results are discussed in Section \ref{expmnt}, followed by the conclusion in Section \ref{con}.

\section{Related Work} \label{Rwrk}
Early detection and classification of  diseases and nutrition deficiencies are inevitable for continuous growth of crops and plants. Ample ML/DL techniques have been developed for plant disease detection and classification. Common crops \textit{e.g.}, potato, rice,  tomato, maize, corn, and many more have been studied in prior works. 

The MobInc-Net  combined MobileNet with the Inception blocks for detecting rice diseases  \cite{chen2022lightweight}. Similarly, a CNN comprising  Inception and residual architecture with a convolution block attention module was discussed \cite{zhao2022ric}. Convolutional autoencoder (CAE) and CNN  were developed for bacterial spot  detection in peach leaves \cite{bedi2021plant}. The CAE was used for dimensionality reduction of  training parameters, and the output of CAE was fed as input to a CNN for classification. 
 Recognition of multiple diseases of maize crop  \textit{e.g.}, anthracnose, rust, etc.  was tested by NPNet-19 \cite{nagaraju2022maize}. An attention-driven method has evaluated multiple public datasets including visual and thermal images of crops at diverse imaging conditions \cite{bera2024attention}.

Using self-attention, a feature aggregation scheme  was  tested for classifying crop diseases  \cite{zuo2022multi}. A generalized region-based attention mechanism described plant disease and pest classification using benchmark datasets and achieved state-of-the-art (SOTA) performances \cite{bera2024rafa}. A shuffle attention method tackled  tomato leaf disease recognition \cite{zhang2023ibsa_net}.

A lightweight CNN  for leaf disease recognition was  evaluated  on  five datasets  \cite{thakur2022vgg}.  A lightweight  fusion strategy with a coordinate attention network  was developed  \cite{chen2022dfcanet}. 
A lightweight attention-based multi-scale feature fusion model employed depth-wise separable convolution for feature extraction and  improved by InceptRes module further, exploiting dual attention module \cite{deng2025damslnet}.
Another lightweight model based on  ResNet architecture  using depth-wise separable convolution  and attention mechanism was developed for classifying  diseases and nutritional deficiencies in corn leaves \cite{timilsina2025cndd}. 
Similarly, a study on predicting plant diseases and nutrient deficiencies of several plants and crops have been identified using a graph convolutional network \cite{bera2024pnd}. A modified pyramid attention network used a dynamic attention mechanism to extract multi-scale features for classifying micronutrient deficiencies (m-ND), severity, and diseases in banana and coffee crops \cite{muthusamy2025dynamic}. 

\begin{figure*}
\includegraphics[width=0.97\textwidth]{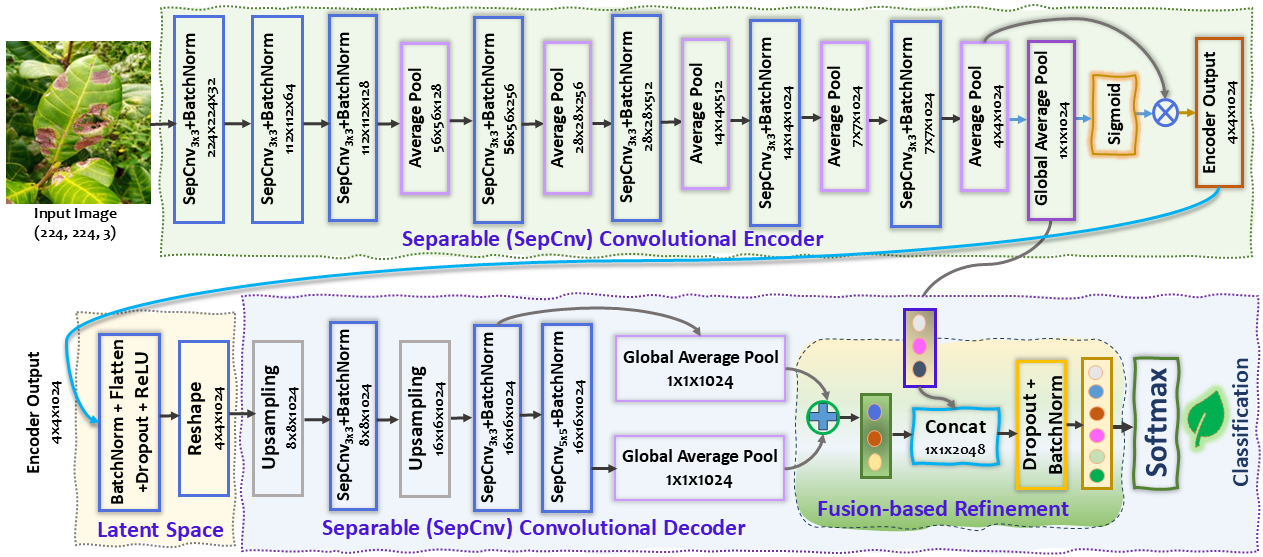} 
\caption{ Diagram of the proposed  convolutional lightweight encoder-decoder model for plant health classification, named CLAP.
}
\label{CLAP_full}
\end{figure*}

A dataset for disease classification of groundnut plants  was introduced in \cite{aishwarya2023dataset}.
A transfer learning approach was developed for   disease classification of groundnut using  pretrained CNN models, namely, ResNet50, InceptionV3, and
DenseNet201 \cite{sasmal2025groundnut}. A similar work attained 96.51\% accuracy on groundnut dataset \cite{sasmal2024novel}. A graph convolutional network was also tested on this dataset \cite{bera2023graph}.

A hybrid vision transformer, named MobilePlantViT, was proposed for generalizing disease classification which optimized resource efficiency  maintaining high performance \cite{tonmoy2025mobileplantvit}. This work achieved good performances on the Cashew, Cassava, Maize,  and Tomato, called CCMT dataset \cite{mensah2023ccmt}.  
A Swin transformer captured global features through shifted windows and self-attention, while,  a dual-attention multi-scale network  combined multi-separable attention  and tri-shuffle convolution attention  block  \cite{karthik2024deep}.  A  CSWinTransformer  enriched global features via  a  feature pyramid   network to refine local feature maps \cite{sun2025efficient}. PWDViTNet has  developed lightweight model for pine wilt disease detection that fused
 ViT and CNN \cite{chen2025pwdvitnet}.

A  Moth-Flame optimized  recurrent neural network extracted the features using the gray level co-occurrence matrix for crop disease and pest detection of CCMT dataset \cite{patel2024optimized}.
A  Multi-kernel inception network was developed for  feature processing, aggregation and diffusion of multi-scale features across hierarchical levels, and another similar module extracted multi-scale  features for enhancing  fusion based feature representation \cite{sun2024multi}.

Several other DL models and transfer learning approaches were used for  evaluating  nutrient deficiencies in chili \cite{bahtiar2020deep}, banana \cite{muthusamy2024incepv3dense}, other plants and crops. 
Most of these prior methods utilized existing backbone CNNs, \textit{e.g.}, ResNet, VGG,  and other lightweight deep architectures. Motivated with such progress, this work develops a  fast lightweight convolutional autoencoder for disease classification of multiple public datasets, consisting of several plants and crops.

\section{Proposed Method: CLAP} \label{method}
We propose  a  convolutional lightweight encoder-decoder  utilizing  depthwise separable convolution as its building block, named CLAP, shown in Fig. \ref{CLAP_full}, described next.

\subsection{ Convolutional Encoder} \label{Enc}
A convolutional encoder extracts the features from an input image $I_k$ $\in$ $\mathbb{R}^{h\times w\times 3}$ with a class-label $k$.   Commonly used standard  convolution is computationally cumbersome requiring  huge model parameters \textit{e.g.}, VGG family. In contrast,  depth-wise  separable convolution (\textit{SepCnv}) is advantageous as it decouples a convolution into depth-wise and point-wise convolutions. We  develop an encoder followed by a decoder using  $\textit{SepCnv}$ with a kernel-size of ${3\times 3}$ that implies a categorical lightweight deep model.   
The  output feature vector of encoder is indicated as      
$\mathbf{F_{En}}$ $\in$ $\mathbb{R}^{h\times w\times C}$ where $h$, $w$, and $C$ imply the height, width, and channels, respectively. The weight is $\mathcal{W}^{(l)}$  and the bias is $b^l$ at $l^{th}$ layer.

\begin{equation} \label{eq0} 
\begin{split}
\mathbf{F}^{l+1}_{En}= BatchNorm\big(ReLU(\textit{SepCnv}(\mathbf{F}^{l})\mathcal{W}^{l}+b^l) \big) 
\end{split}
\end{equation}

The encoder consists of multiple  \textit{SepCnv}  blocks with increasing channel dimensions.  The filter-size of each convolutional layer $l$ is given as $[32,64,128, 256, 512, 1024]$.  
ReLU activation for non-linearity and Batch Norm for regularization are included in encoder, followed by an average pooling for essential feature selection via downsampling  spatial dimension.
\begin{equation} \label{eq1} 
\begin{split}
\mathbf{F}^{l+1}_{En^{(pool)}}=\textit{Dropout} \big(\textit{AveragePool}(\mathbf{F}^{l}) \big) 
\end{split}
\end{equation}
\begin{equation} \label{eq4} 
\mathbf{\tilde{F}}_{En}=\sigma\big(\textit{GAP}\big(\mathbf{F}_{En^{(pool)}} \big)\big) \otimes\mathbf{F}_{En^{(pool)}} 
\end{equation}
A dropout layer is applied to overcome overfitting that is a common challenge in DL. A dropout scheme indeed enhances the training performance by discarding  features randomly.  A commonly used gating-based attention method is applied using a sigmoid ($\sigma$) function. Next, the down-sampled feature map, $ F_{En}^ {4\times 4\times 1024}$, is squeezed through a global average pooling (GAP) layer followed by a sigmoid activation. This transformed feature map is then multiplied with the actual  feature map $F_{En}$ to produce a gated-output of encoder block, denoted as $\tilde{F}_{En}$ keeping the same feature map dimension ${4\times 4\times 1024}$. It acts as a simple attention-gating for refining to enrich encoder's representational capacity. 
The tiny encoder consists of only a few number of convolutional layers adhering a small parametric count, offering for a shallower design, hence, a lightweight encoder. Next, the latent feature map ($\mathbf{{F}}_{Lt}$) is represented by a flatten layer, followed by a ReLU non-linear activation  and reshaped the feature map suitably as an input of decoder module.

\begin{equation} \label{eq4} 
\mathbf{{F}}_{Lt}=Reshape \big(ReLU \big(\textit{Flatten}\big(\mathbf{\tilde{F}}_{En} \big)\big) \big)
\end{equation}

\subsection{ Convolutional Decoder} \label{Dec}
The decoder block is built with upsampling the latent feature map followed by similar $SepCnv$ layers, and repeated twice. This produces a feature map of size $8\times 8\times 1024$, then, $16\times 16\times 1024$, respectively. The output of decoder is given as

\begin{equation} \label{eq7} 
\begin{split}
\mathbf{F}^{(l+1)}_{De}=\textit{Dropout}\big( \textit{SepCnv}\big( \textit{Upsampling}(\mathbf{F}_{Lt}^{l}) \big) \big). 
\end{split}
\end{equation}
Again, a convolution layer with a larger receptive field size with a $5\times5$ kernel is applied for capturing larger semantics of feature map. Then, two GAP layers are used to select the essential features rendered from decoder layers of two different receptive field sizes, and added them for mixing their discriminability in capturing complex visual patterns.
\begin{equation} \label{eq6} 
\begin{split}
\mathbf{\tilde{F}}_{De}=\textit{GAP}\big( \mathbf{F}_{De{(3\times3)}}\big)+ \textit{GAP}\big(\mathbf{F}_{De{(5\times5)}}\big)
\end{split}
\end{equation}
Then, the pooled output of decoder block is combined with the pooled encoder's feature map obtained after the last convolutional layer as a residual path to address vanishing gradient problem. A simple fusion of  encoder-decoder modules through concatenating both feature maps enrich overall representational capacity of CAE which is effective for classification. The output feature vector $\mathbf{{F}}_{CAE}$ of the CAE is given as 
\begin{equation} \label{eq7} 
\centering
\begin{split}
\mathbf{{F}}_{CAE}&=\textit{Concat} \big[ \textit{GAP}\big(\mathbf{F}_{En^{(pool)}}\big);  \mathbf{\tilde{F}}_{De}\big]; \\ 
\mathbf{K}_{pred}&={softmax} \big({\mathbf{F}_{CAE}}\big).
\end{split}
\end{equation}

Final prediction is made by passing  $\mathbf{{F}}_{CAE}$ via a softmax layer.
The predicted class-label $\bar{k} \in \textbf{K}_{pred}$, pertaining to  actual class-label $k\in K$. The categorical cross-entropy loss function is used. 
The proposed CLAP contains a total of  4,991,554 parameters($\approx$ 5 million) implying a lightweight convolutional autoencoder (CAE), offers a faster training-testing time.

\section{Experiments and Discussion} \label{expmnt}
First the datasets used for evaluation are precisely specified, then implementation details with results are discussed.

\subsection{Dataset Summary}
\begin{figure}[h]
\centering
\includegraphics[width=0.23\linewidth, height=2.5 cm ]{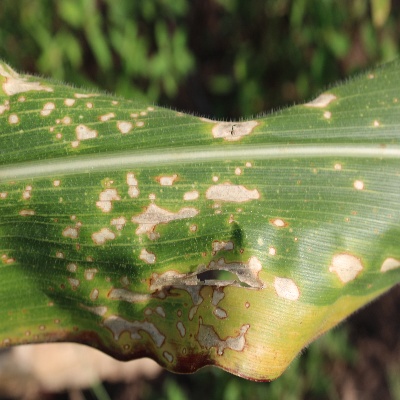}
\includegraphics[width=0.23\linewidth, height=2.5 cm ]{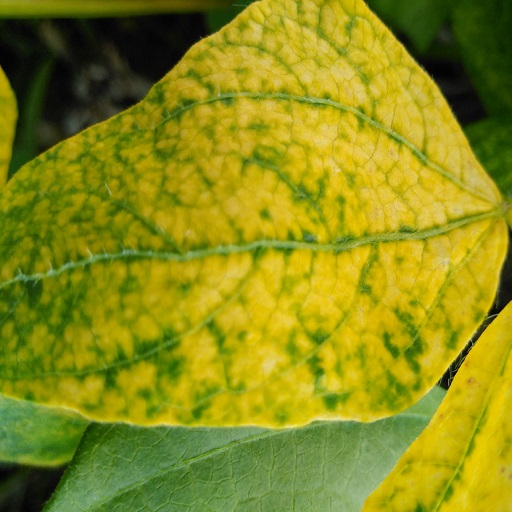}
\includegraphics[width=0.23\linewidth, height=2.5 cm ]{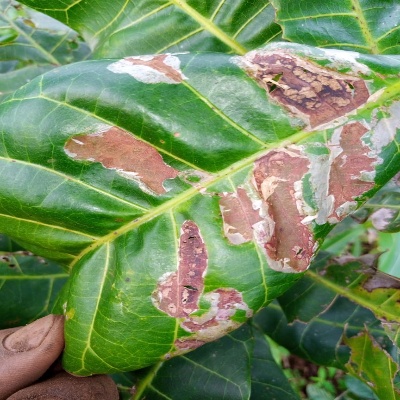}
\includegraphics[width=0.23\linewidth, height=2.5 cm ]{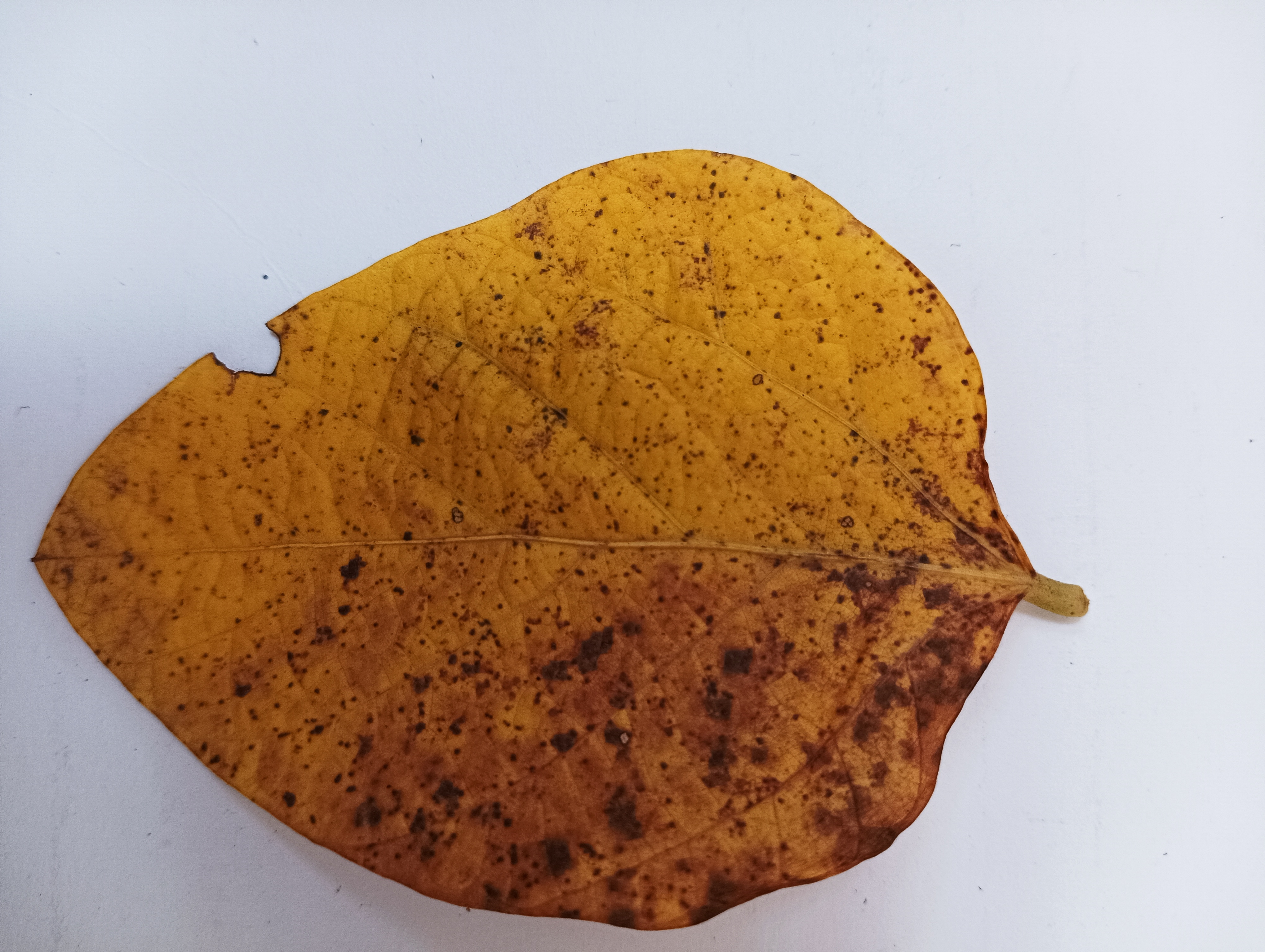}
\caption{{Sample images of disease-infected plants: maize leaf blight, blackgram yellow mosaic,   cashew leaf miner, and soybean septoria brown spot. } 
}
\label{fig:sampleEx}
\end{figure}

(a) \textbf{Integrated Plant Disease (IPD) dataset}
is a collection of multiple  datasets representing diseases and healthy leaves of 22 different plants, which were collected at various imaging constraints mainly from different states of India. Details of the plants and their disease categories are provided in Table \ref{tab:cr}, and samples are illustrated in Fig \ref{fig:sampleEx}.  

We divided IPD dataset into different splits based on each class separately, and finally combined into a bigger diverse dataset. It eliminates scope of redundancy in mixing samples of different splits in a mutually exclusive manner. This heterogeneous IPD dataset is a collection of multi-source datasets which are integrated into a single dataset maintaining a unified split-ratio of each class of plant disease. 

(i) \textbf{Sugarcane Leaf Disease Dataset}\footnote{Daphal, Swapnil; Koli, Sanjay (2022), “Sugarcane Leaf Disease Dataset”, Mendeley Data, V1, doi: 10.17632/9424skmnrk.1}
was  collected manually in Maharashtra, India. It consists of  the healthy, mosaic, redrot, rust and yellow diseases. The images were captured with smartphones  maintaining various configurations. 

(ii) \textbf{Soybean disease dataset}\cite{kotwal2024india} represents  
 five types of disease namely, healthy, vein necrosis, dry leaf, 
brown spot, and bacterial blight. It was acquired in Maharashtra, India. 

 (iii) \textbf{Blackgram Dataset}\footnote{Talasila, Srinivas; Rawal, Kirti; Sethi, Gaurav; MSS, Sanjay; M, Surya Prakash Reddy (2022), “Blackgram Plant Leaf Disease Dataset”, Mendeley Data, V3, doi: 10.17632/zfcv9fmrgv.3}  comprises five disease categories, and was collected in  Andhra Pradesh, India. It consists of common leaf diseases namely the anthracnose, crinkle, powdery mildew, and yellow mosaic.

 (iv) \textbf{Grapes Leaf Disease Dataset}\cite{dharrao2025grapes} is  introduced recently. With healthy leaves, it contains the following disease classes: downy mildew, powdery mildew and bacterial spot.  

 (v) \textbf{Betel Leaf Disease Dataset}\cite{mane2025pribel}  was collected from Pune, India, under natural and controlled conditions. Here, the natural imaging scenario is considered for evaluation. It includes 
healthy, diseased, and dried classes of leaves.  

Altogether,  total  8327 images are used in our experiments, detailed in Table \ref{tab:cr}. 
The dataset is divided into 60\% train (5033), 20\% validation (1647), and remaining 20\% for testing.

(b) \textbf{Groundnut Dataset} \cite{aishwarya2023dataset}
 represents groundnut plant leaves collected  in Koppal, Karnataka, India.  The images are categorized into 6 classes:  healthy leaves,  early leaf spot, late leaf spot, nutrition deficiency, rust and early rust (Table \ref{DB2}).

(c) \textbf{CCMT Dataset} \cite{mensah2023ccmt}
 consists of  high-quality images of four crops namely, cashew, cassava, maize, and tomato. Overall, it has 22 different classes representing plant leaves, pests, fruits and images of sick parts of plants (Table \ref{DB2}).

\begin{table}[!htp]
\centering
\caption{Ablation study: Module-wise Results on IPD Dataset}\label{CLAP2}
\begin{tabular}{c c c c  c  c} 
 \hline
  Method & Val Acc  & Test Acc & Prec   & Rec & F1-sc \\           \hline

  Base Encoder  & 95.33 & 93.29 &93.0   & 93.0 & 93.0\\

 Encoder+Decoder-I  &95.85  &94.33  &94.0   & 94.0 & 94.0\\

\textbf{CLAP}  & \textbf{96.03} & \textbf{95.67} &\textbf{ 96.0}   & \textbf{96.0} & \textbf{96.0}\\

\hline
\end{tabular}
\end{table}

\begin{table}[!htp]
\centering
\caption{Summary of the Integrated Plant Disease (IPD) Dataset and class-wise overall performance of CLAP } \label{tab:cr}
\begin{tabular}{c c   c c c  c} 
 \hline
 Plant & Train:Test  & Class &  Prec   & Rec & F1 \\           \hline


      &  &  Diseased  &     89.0   &  89.0  &    89.0       \\
       Betel &553:177 & Leaf Dried   &    93.0     & 93.0   &   93.0      \\
   &   &  Healthy    &    97.0    &  95.0   &   96.0        \\ \hline
     
  &     & Anthracnose   &    96.0  &    96.0   &   96.0    \\
  &    & Healthy   &    94.0    &  77.0  &    85.5      \\
      Blackgram & 605:201 & Leaf Crinkle  &    83.0   &   100.0  &    91.5    \\
  &    & Powdery Mildew  &     88.0  &    97.0  &    92.5     \\
&    & Yellow Mosaic  &     94.0    &  98.0  &    96.0      \\ \hline

&     & Bacterial Spot &     87.0   &   100.0  &    93.5      \\
&& Downy Mildew  &     100.0   &   99.0 &     99.5      \\
      Grapes  &  1638:544 & Healthy Leaves  &     99.0    &  100.0    &  99.5      \\
 &     & Powdery Mildew  &     99.0 &     98.0  &    98.5      \\ \hline
 
 &     & Healthy  &     100.0   &   100.0 &     100.0      \\
  &     & Vein Necrosis  &     96.0   &   100.0  &    98.0   \\
       Soybean & 718:224 & Dryleaf &     100.0     & 100.0    &  100.0     \\
 & & Septoria Brown    &    100.0   &   100.0  &    100.0    \\
 & &  Leaf Blight &     98.0  &    100.0    &  99.0     \\ \hline
 &       & Healthy   &    96.0   &   92.0 &     94.0   \\
 &     &Mosaic &     88.0   &   93.0   &   90.5   \\
        Sugarcane &1520:500 & RedRot &     96.0   &   93.0 &     94.5    \\
  &      & Rust   &    92.0 &     94.0   &   93.0  \\
  &     & Yellow &     94.0  &    86.0  &    90.0   \\
\hline
Weighted  & Average &  22 classes &   96.0    &  96.0  &    96.0  \\

\hline
\end{tabular}
\end{table}

\begin{table}[h]
\caption{Left: Dataset specifications (No. Classes: \#C) with train-test sizes. Middle: Overall Accuracy (Acc) of MobileNet(MbN) and CLAP. Right: CLAP's performances with other metrics (\%) } \label{DB2}
\begin{center}
\begin{tabular}{c c |c c| ccc}
\hline 
Dataset\#C & Train : Test & MbN & CLAP & Pre & Rec & F1 \\
\hline 
IPD-22 & 5033:1647 & \underline{95.62}  & \textbf{95.67} &96.0   & 96.0 & 96.0\\

Groundnut-6  & 7910:2451 & \underline{95.54} & \textbf{96.85} & 97.0& 97.0 &97.0  \\ \hline
Cassava-5 & 5630:940 &93.48 &94.02 & 94.0& 94.0 &94.0\\
Cashew-5 & 4906:826 & 94.11& 94.00 & 94.0& 94.0 &94.0\\
Maize-7 & 3972:663  &80.33  & 82.77 & 83.0& 83.0 & 83.0\\
Tomato-5 & 4338:721 & 73.61& 76.66 & 77.0& 77.0 & 77.0\\
\hline
CCMT-22 & 18846:4150 &\textbf{87.28}  &  \underline{87.11} & 87.0 & 87.0 & 87.0 \\
\hline
\end{tabular}
\label{ccmt}
\end{center}
\end{table}

\subsection {Implementation Specification} \label{implemnt}
 With the basic pre-processing methods of Keras applications,  image augmentation such as  rotation ($\pm$25 degrees), scaling ($\pm$0.25), and cropping with size 224$\times$224 from input size 256$\times$256 are applied for data variations. The CLAP is implemented in Tensorflow 2.x using Python and trained for 300 epochs, starting with a learning rate of $0.008$. The model parameters are computed in million (M), and model complexity is estimated in Giga FLOPs (GFLOPs), and speed of training-testing in milliseconds (ms). 

The   accuracy, precision, recall, and F1-score are used as standard evaluation metrics which are  widely used for evaluation  even though samples of various classes are imbalanced. The performance is also measured  using confusion matrix implying a reliable assessment. 



\begin{equation} \label{metric}
  \begin{split}
{Accuracy}&=\frac{TP+TN}{TP+TN+FP+FN}\ \\  
 {Precision }&=\frac{TP}{TP+FP}\  \\
 {Recall} &= \frac{TP}{TP+FN}\  \\
 {F1}\text{-}{Score }&= 2\times\frac{{Precision}\times {Recall}}{{Precision} + {Recall}}  
 \end{split} 
 \end{equation}
  
\noindent where TP is the number of true positives, TN is the number of true negatives, FP is the number of false positives, and FN denotes the number of false negatives.

\begin{figure}[h]
\centering
\includegraphics[width=0.9\linewidth, height=6.0 cm ]{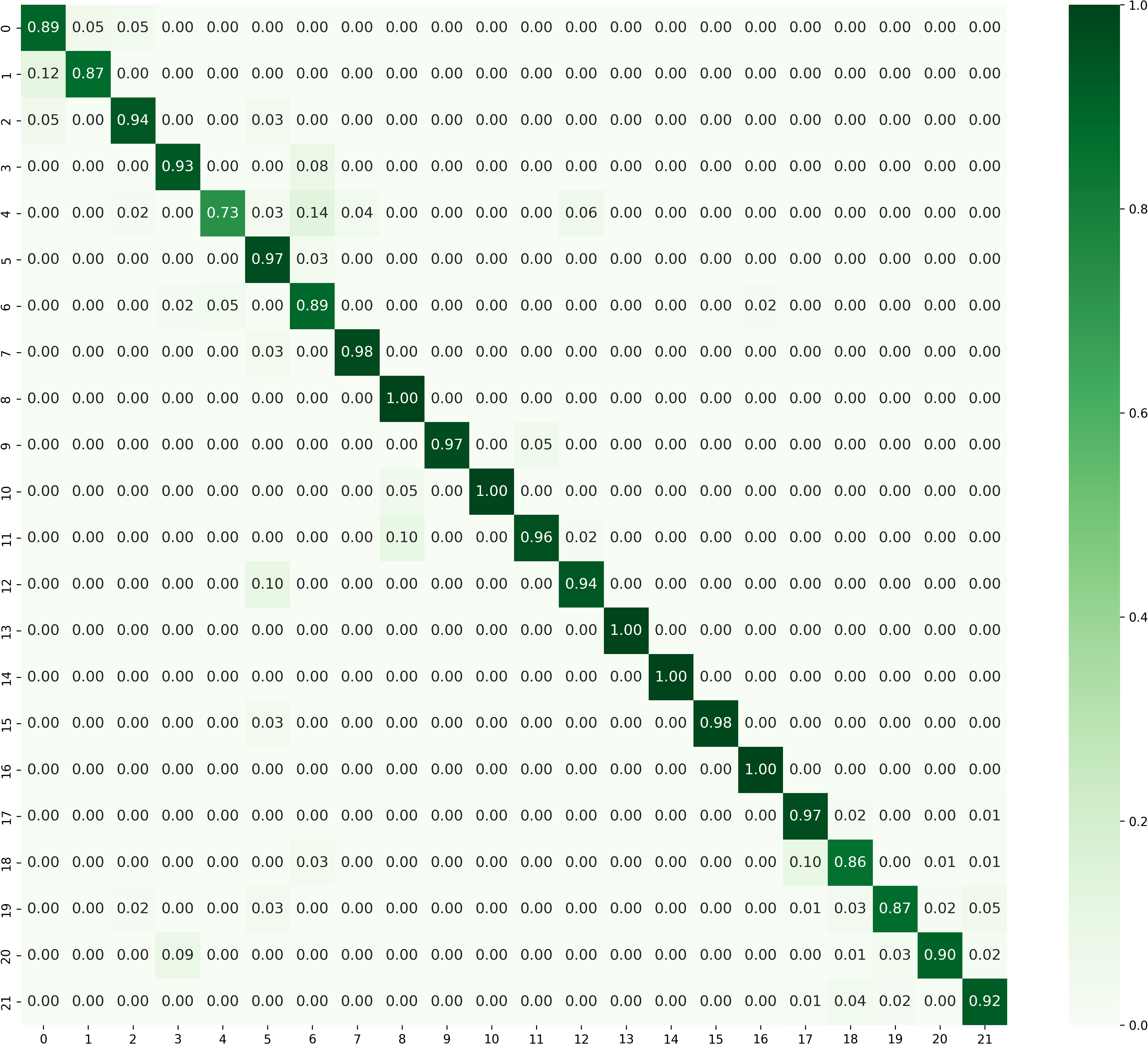}
\caption{{Confusion matrix of CLAP on the IPD dataset} 
}
\label{fig:cm1}
\end{figure}

\subsection{Performance Analysis}
In our IPD dataset,  experiments are conducted with all 22 classes. In contrast, experiments on  CCMT are two folds, \textit{i.e.}, separate experiments on each sub-datasets as well as on all CCMT sub-classes together in another experiment. The reason of both types of experiments is to compare with SOTA. Because, a few works have studied plant-wise experiments, whereas others tested on full dataset.

The  CLAP has attained 95.67\% accuracy on the IPD dataset, and this accuracy is near to MobileNetV2's baseline result (95.62\%), which is trained from scratch. The results are given in Table \ref{CLAP2}-\ref{DB2}.  Whereas, only the base encoder is effective in attaining 93.29\% accuracy, implying the aptness of decoder with attention gating for further feature refinement to enhance performance. Also, the encoder and  with only one upsampling layer in decoder i.e., Decoder-I ($8\times 8 \times 1024$) provides 94.33\% test accuracy. Thus, refinement by mixing feature with more decoding layer indeed beneficial for  performance improvement.  The confusion matrices are shown in Fig. \ref{fig:cm1}. The results are given in Table \ref{CLAP2}. An  in-depth result showing class-wise performance on IPD dataset is given in Table \ref{tab:cr}.

\begin{figure}[h]
\centering
\includegraphics[width=0.46\linewidth, height=3.3 cm ]{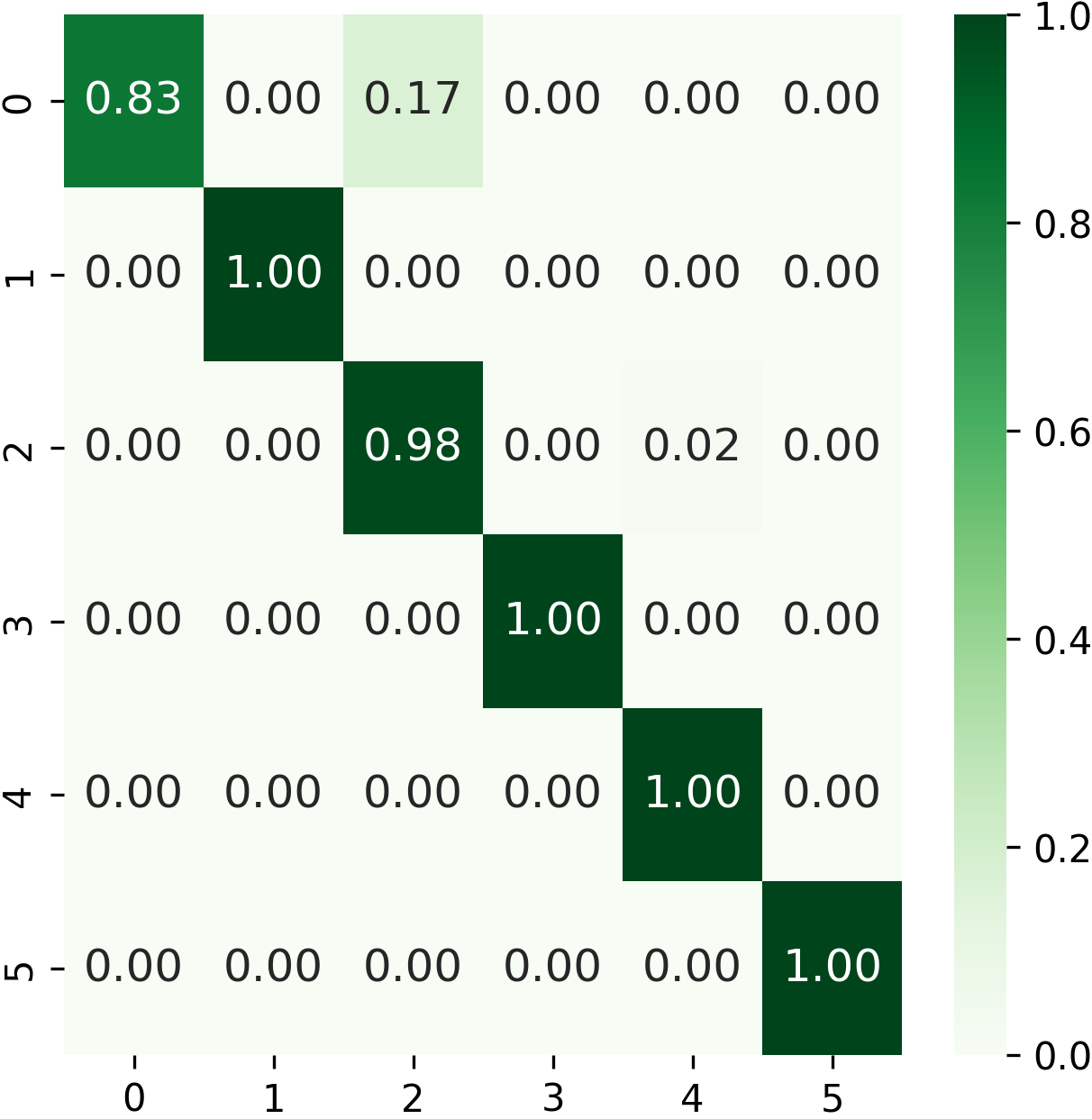} \hfill
\includegraphics[width=0.46\linewidth, height=3.3 cm ]{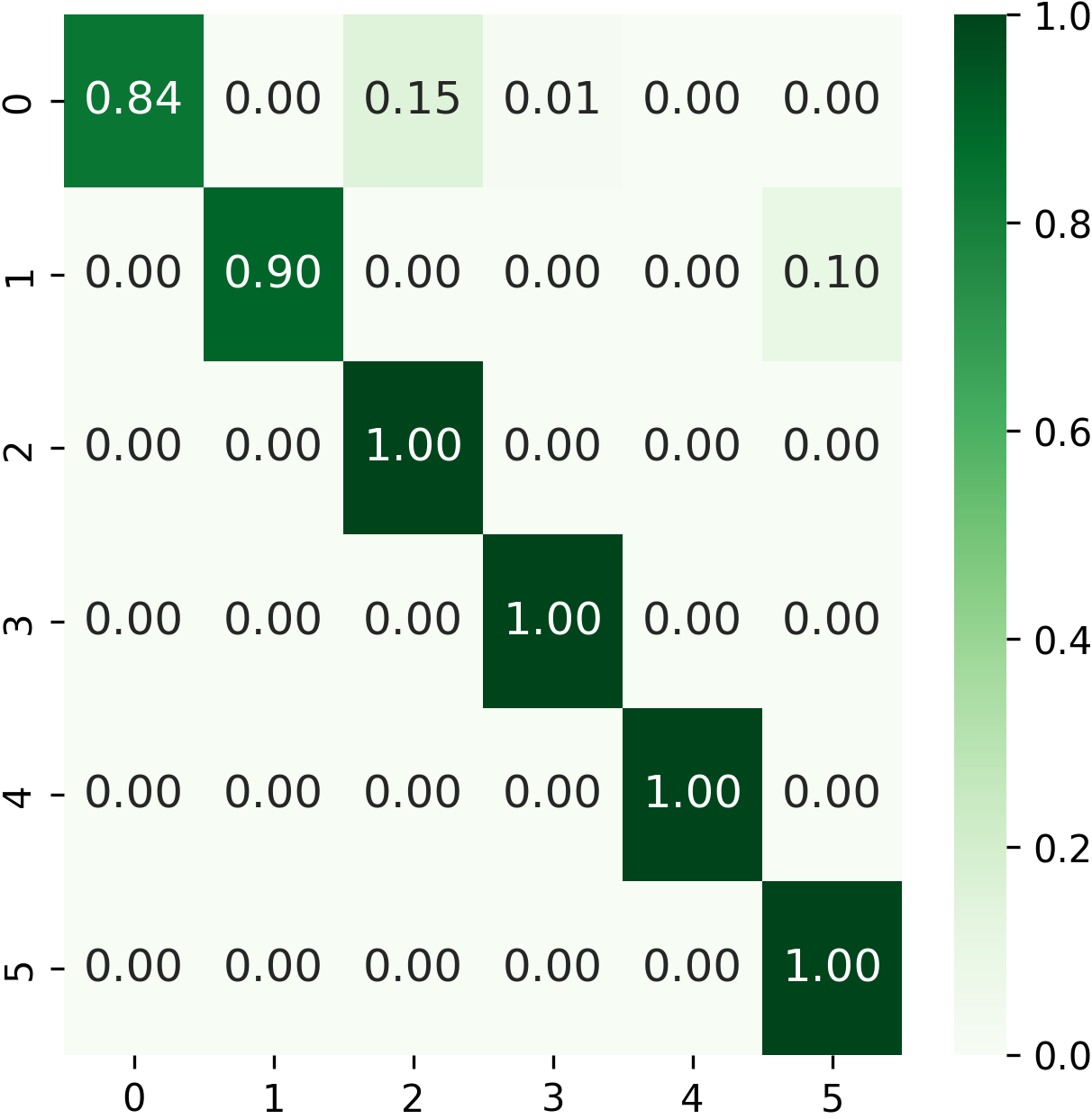}
\caption{{Confusion matrix on Groundnut, left: CLAP, right:MobileNetV2 } 
}
\label{fig:GN_cnmtx}
\end{figure}

The  CLAP has gained 96.85\% accuracy on the  groundnut dataset. In contrast, MobileNetV2-based method reported 95.60\% accuracy \cite{paramanandham2024enhancing};  Xception gained 93.71\%  \cite{aishwarya2023ensemble}, and 96.70\% \cite{bera2023graph}  accuracies on this dataset. Also, higher performances achieved using pre-trained heavier models like ResNet, DenseNet, and others \cite{sasmal2025groundnut}. However, those heavier models are not comparable to our approach due to different experimental constraints.
Class-wise performance on this dataset is provided in Table \ref{Gnut}, where the results of CLAP and MobileNetV2 are shown in the left- and right-side column-sets, respectively. The confusion matrices of this study are shown in Fig. \ref{fig:GN_cnmtx}.

\begin{table}[h]
\caption{Performance on Groundnut, left: CLAP; right: MobileNet} \label{Gnut}
\begin{center}
\begin{tabular}{c c c c| c c c }
\hline 

     Class name        & Prec &   Rec & F1-sc  & Prec  & Rec &  F1-sc  \\
\hline
           Early leaf spot &      100.0    &   82.0    &  91.0   &   100.0   &      84.0    &   92.0  \\
           Early rust  &    100.0    &   100.0   &    100.0    &    100.0    &     90.0  &     95.0  \\
           Healthy leaf  &     85.0   &   98.0   &   91.5   &    87.0   &     100.0    &    93.5      \\
         Late leaf spot   &   100.0     &  100.0    &   100.0    &   99.0   &    100.0  &       99.5    \\
          Nutrition deficit   &   100.0     &  100.0     &  100.0     &    100.0    &     100.0  &     100.0    \\
           Rust   &   100.0    &   100.0   &    100.0    &  91.0    &  100.0     &  95.5  \\
\hline

Weighted avg  &     97.0   &   97.0   &   97.0  &  96.0   &    96.0    &   96.0 \\

\hline
\end{tabular}
\label{gnut}
\end{center}
\end{table}

\begin{figure}[h]
\centering
\includegraphics[width=0.45\linewidth, height=3.3 cm ]{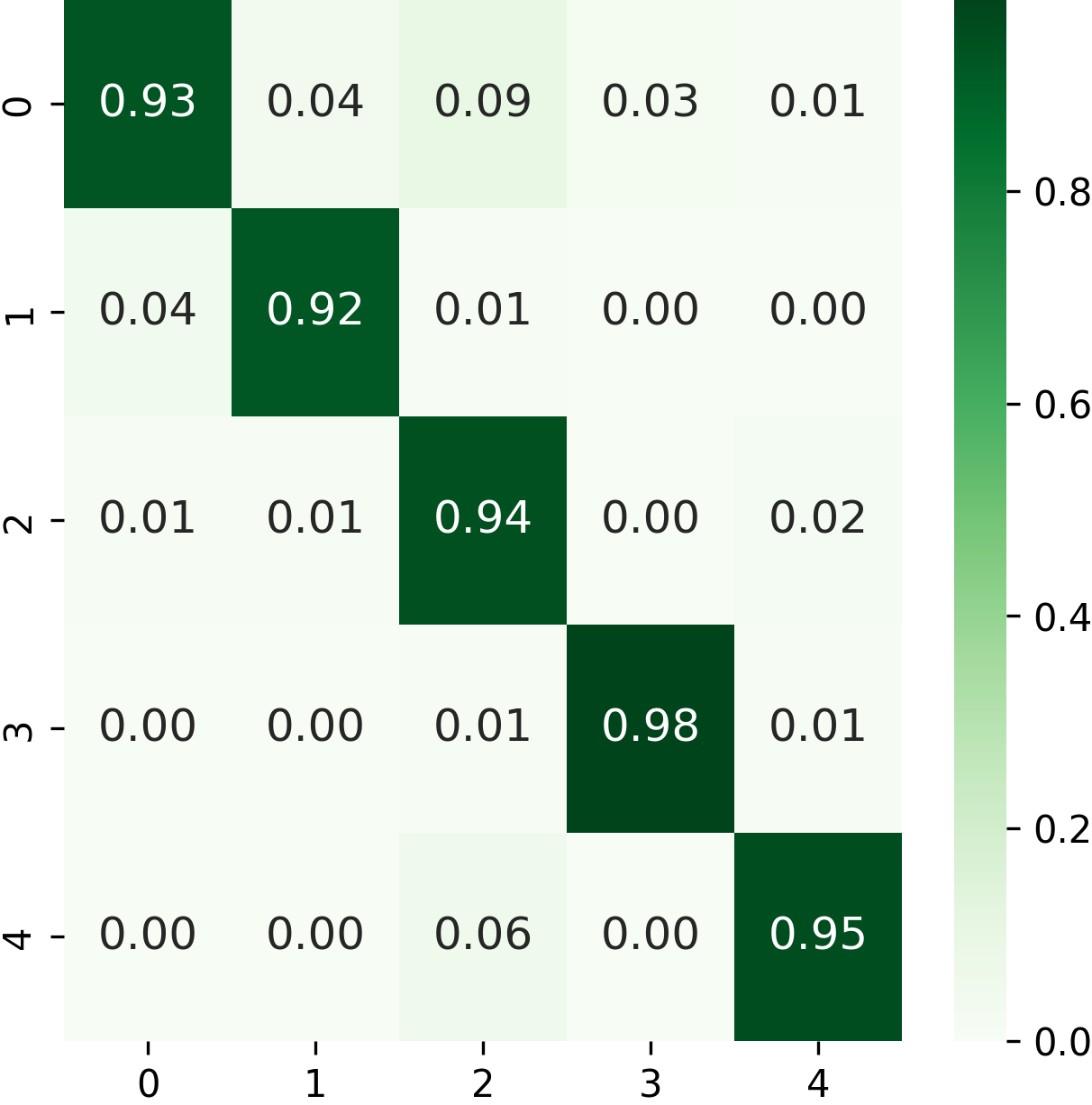} \hfill
\includegraphics[width=0.45\linewidth, height=3.3 cm ]{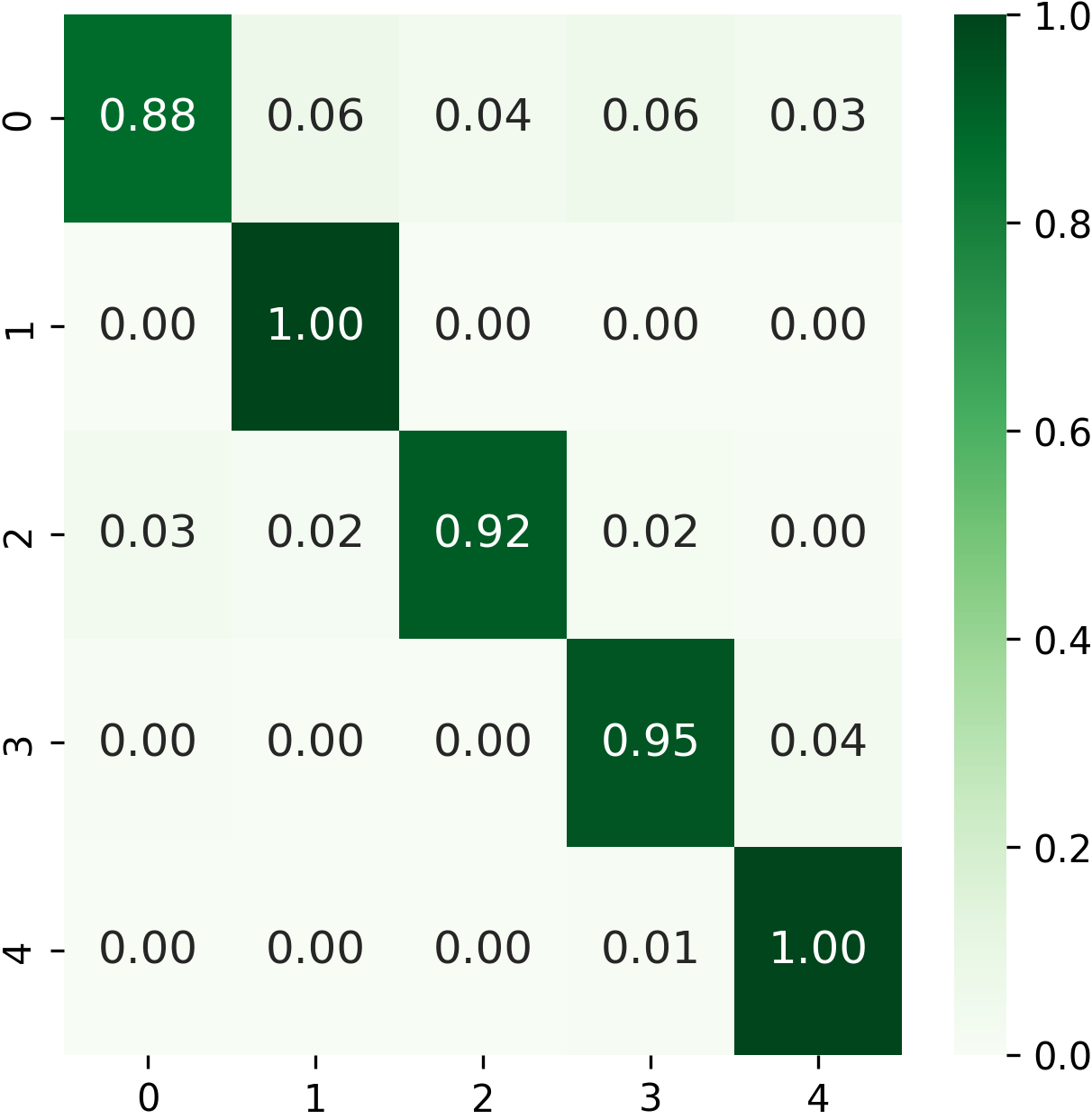} \hfill
\\
\vspace{0.4cm}

\includegraphics[width=0.45\linewidth, height=3.30 cm ]{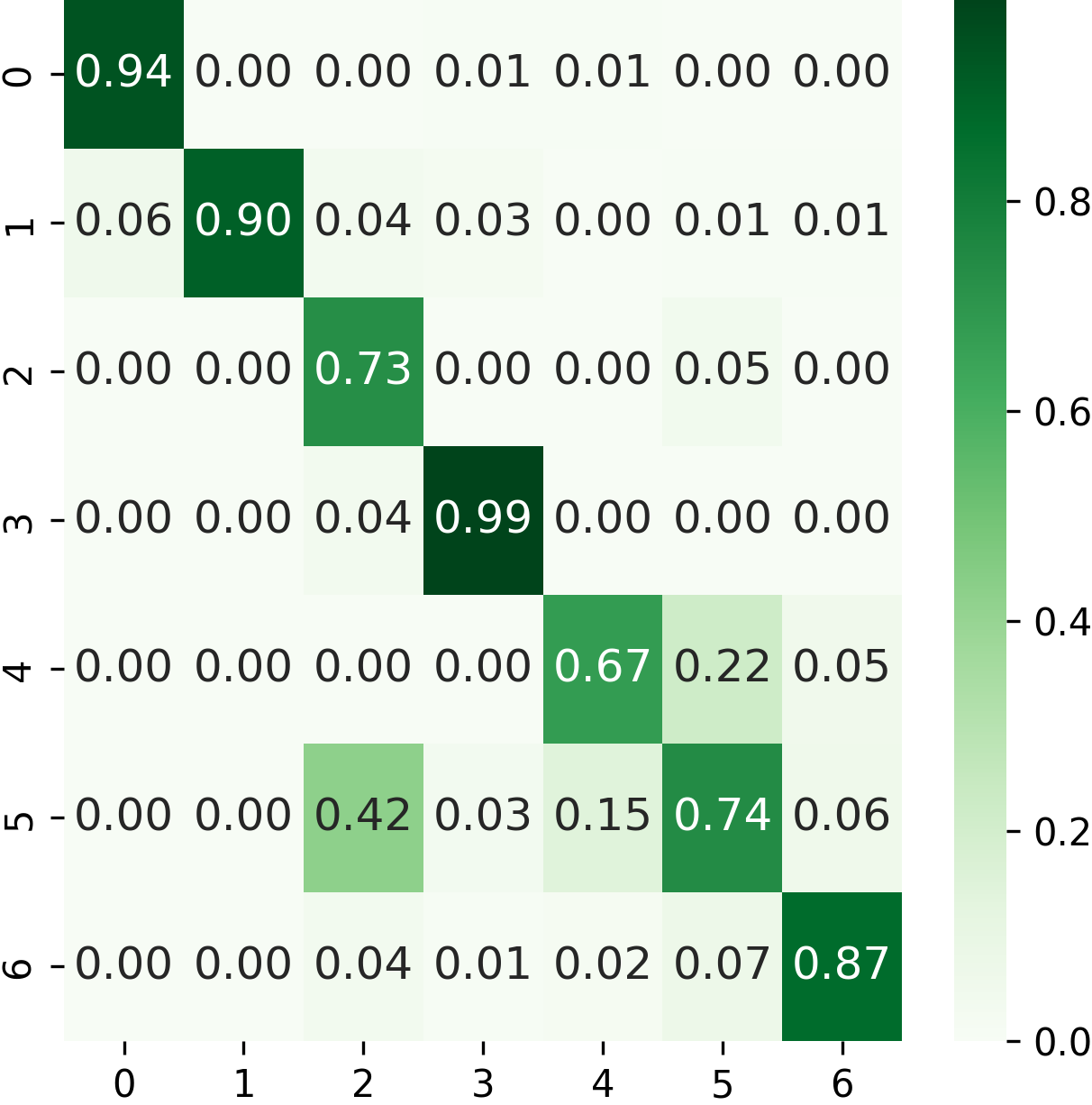} \hfill
\includegraphics[width=0.45\linewidth, height=3.30 cm ]{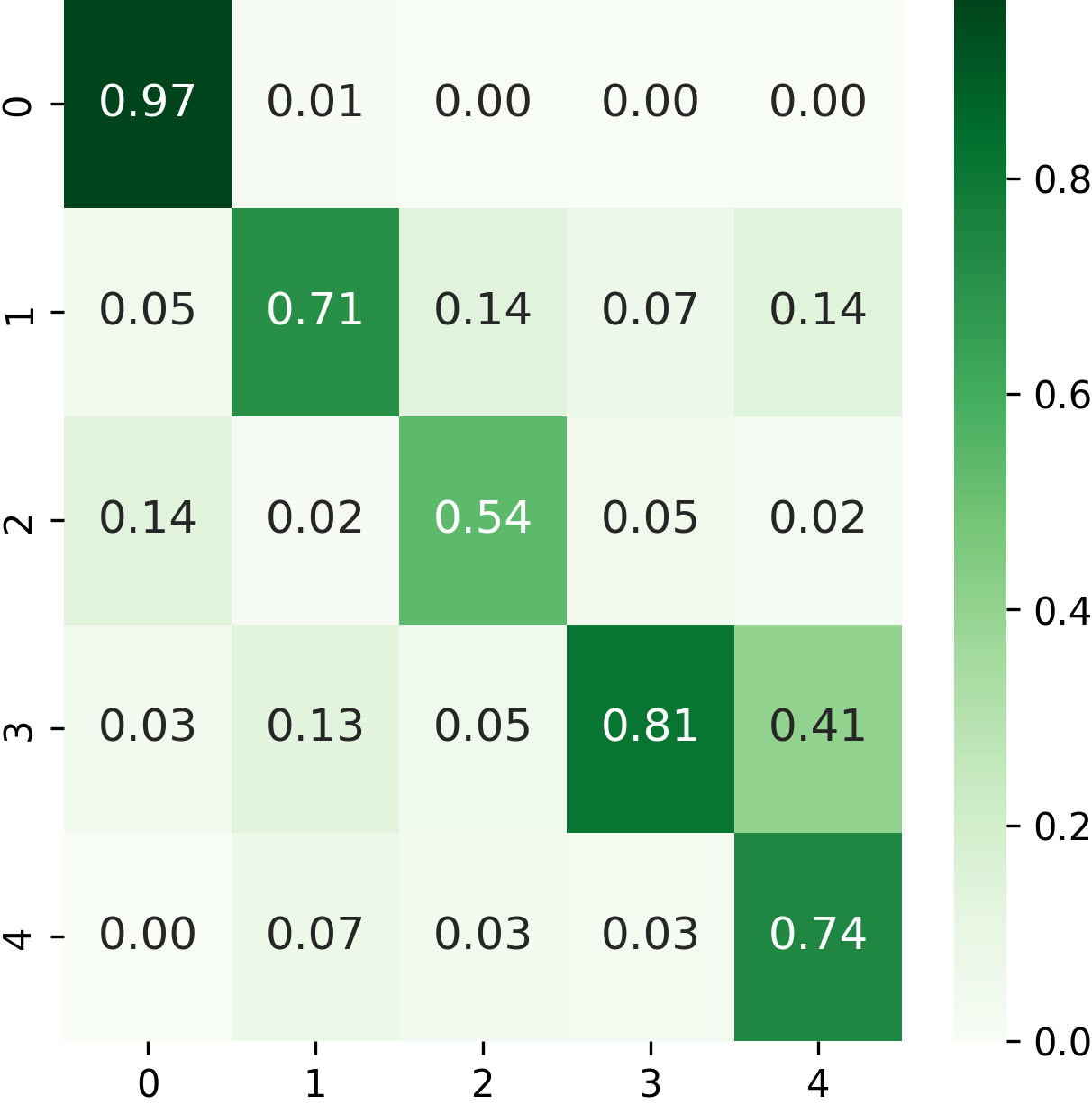} \hfill
\caption{{Confusion matrix on  CCMT (sequence given in Table III) dataset} 
}
\label{fig:CCMT_cnmtx}
\end{figure}

Transfer learning with pretrained MobileNetV2 achieved 76.78\% accuracy on tomato plant (CCMT) diseases  \cite{bahrami2024tomato}. MobilePlantViT \cite{tonmoy2025mobileplantvit} achieved an accuracy of 95.04\% on cashew, 94.34\% on cassava, 81.05\% on maize, and  80.05\% on tomato leaves of CCMT, respectively. In contrast, CLAP has attained 94.0\% on cashew, 94.02\% on cassava, 82.77\% on maize and 76.66\% on tomato plants respectively. Notably, the performances of CLAP is competitive with lightweight MobileNetV2 baselines (87.28\%), in Table \ref{DB2}. The confusion matrices of each plants are shown in \ref{fig:CCMT_cnmtx}. Thus, CLAP has gained improved or competitive results aligned with MobileNetV2's baseline performance on all aforesaid datasets.


\begin{figure}[h]
\centering
\includegraphics[width=0.3\linewidth, height=2.5 cm ]{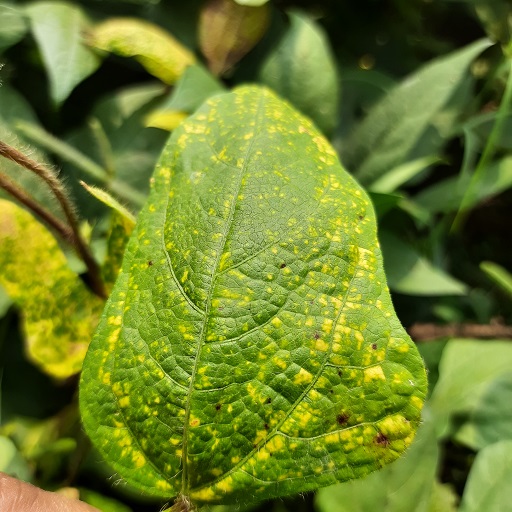}
\includegraphics[width=0.3\linewidth, height=2.5cm ]{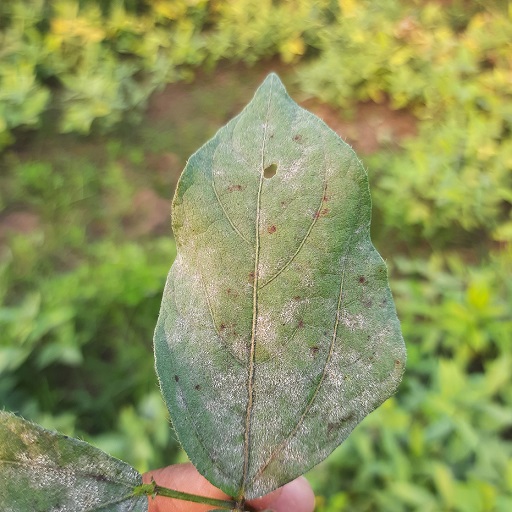}
\includegraphics[width=0.3\linewidth, height=2.5 cm ]{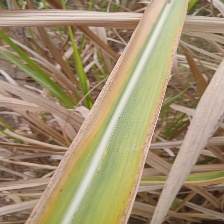}\\
\vspace{0.2 cm}
\includegraphics[width=0.3\linewidth, height=2.5 cm ]{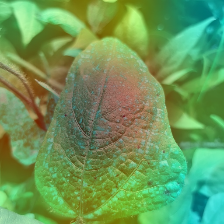}
\includegraphics[width=0.3\linewidth, height=2.5 cm ]{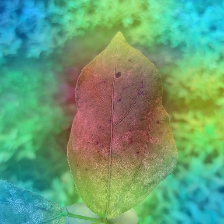}
\includegraphics[width=0.3\linewidth, height=2.5 cm ]{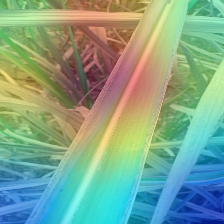}
\caption{{Grad-CAM images  show  vital regions of different infected leaves} 
}
\label{fig:Grad-CAM}
\end{figure}

{\textbf{Model Complexity:} The average training time of CLAP on CCMT-Maize (3972 images) is 77 sec (20 ms/img) and inference time is $\approx$1 ms/img. The training time is 95 sec ($\approx$24 ms/img) and 2.2 ms/img for inference by MobileNetV2.  The GFLOPs of CLAP is 0.2, comparatively better than 0.6 of MobileNetV2.} Faster training and inferencing of CLAP implies its suitability for real-time deployment.

\textbf{Visualization:}
The confusion matrices on aforesaid  plant datasets (given in Fig. 4-6) show overall performances of CLAP.  The gradient-weighted class activation mapping (Grad-CAM)  is commonly used for visual explanations of class-wise discriminability to indicate the efficacy of a model \cite{selvaraju2017grad}. Examples are shown in Fig. \ref{fig:Grad-CAM}  for visual explainability which showcase  vital infected part of a leaf captured by the CLAP. 

\vspace{0.2 cm}
\section{Conclusion} \label{con}
This work has presented  a convolutional lightweight autoencoder for plant disease classification, named CLAP. The CLAP is a simple integration of a few convolutional layers following an encoder-decoder architecture, and finally fuses the features of encoder-decoder to enrich overall feature discriminability.
The tiny CLAP  requires only 5 M parameters and 0.2 GFLOPs, compared to existing heavier models. The performances have been computed  on several recent public datasets representing diverse plant leaf diseases.  CLAP has improved the performances on various datasets as described above. Moreover, the results are very close to  lightweight MobileNetV2 backbone. The training and inferencing of CLAP is faster, implying its suitability for real-time deployment. In the future, new deep learning methods like autoencoders will be developed for plant disease classification.  We encourage the researchers  for improvement towards  early-stage detection of abnormalities, essential nutrition and soil factors, vital for sustainable agricultural growth.
\vspace{0.2 cm}
\section*{Acknowledgment}
{ This work was supported by the ANRF PMECRG, File No: ANRF/ECRG/2024/001743/ENS. The authors are thankful to BITS Pilani, India for providing computational infrastructure. }

\vspace{0.4 cm}
\bibliographystyle{IEEEtran}
\bibliography{IEEEexample.bib}

\end{document}